\documentclass[10pt, a4paper]{article}

\usepackage{lrec-coling2024} 
\usepackage{multirow,diagbox,makecell,amssymb,bm}
\usepackage{amsmath,amsthm}
\usepackage{makecell}
\usepackage{graphicx}
\usepackage{pifont}
\usepackage{algorithm}
\usepackage{algorithmic}
\usepackage{amsfonts}       
\usepackage{booktabs}       
\usepackage{footmisc}

\title{
 How Good Are LLMs at Out-of-Distribution Detection?
 }

\name{Bo Liu$^{1*}$\thanks{~$^*$ Equal contribution.}, Li-Ming Zhan$^{1*}$, Zexin Lu$^{1*}$, Yujie Feng$^1$, Lei Xue$^2$, Xiao-Ming Wu$^{1\dag}$\thanks{~$^\dag$ Corresponding author.}} 

\address{Department of Computing, The Hong Kong Polytechnic University, Hong Kong S.A.R.$^1$\\ 
School of Cyber Science and Technology, Sun Yat‐Sen University, Shenzhen, China.$^2$ \\
         \{bokelvin.liu, lmzhan.zhan, zexin.lu, yujie.feng\}@connect.polyu.edu.hk\\
         xuelei3@mail.sysu.edu.cn\\
         xiao-ming.wu@polyu.edu.hk\\}

\abstract{
Out-of-distribution (OOD) detection plays a vital role in enhancing the reliability of machine learning models.
As large language models (LLMs) become more prevalent, the applicability of prior research on OOD detection that utilized smaller-scale Transformers such as BERT, RoBERTa, and GPT-2 may be challenged, due to the significant differences in the scale of these models, their pre-training objectives, and the paradigms used for inference.
This paper initiates a pioneering empirical investigation into the OOD detection capabilities of LLMs, focusing on the LLaMA series ranging from 7B to 65B in size. We thoroughly evaluate commonly used OOD detectors, examining their performance in both zero-grad and fine-tuning scenarios. Notably, we alter previous discriminative in-distribution fine-tuning into generative fine-tuning, aligning the pre-training objective of LLMs with downstream tasks.
Our findings unveil that a simple cosine distance OOD detector demonstrates superior efficacy, outperforming other OOD detectors. 
We provide an intriguing explanation for this phenomenon by highlighting the isotropic nature of the embedding spaces of LLMs, which distinctly contrasts with the anisotropic property observed in smaller BERT family models. The new insight enhances our understanding of how LLMs detect OOD data, thereby enhancing their adaptability and reliability in dynamic environments. 
We have released the source code at \url{https://github.com/Awenbocc/LLM-OOD} for other researchers to reproduce our results.
 \\ \newline  }

\begin{document}

\maketitleabstract

\section{Introduction}


Out-of-distribution (OOD) detection has attracted significant attention due to its crucial role in ensuring AI safety~\cite{DBLP:journals/tmlr/SalehiMHLRS22}. The objective is to identify and raise an alarm for inputs that exhibit distributional shifts compared to the in-distribution (ID) training data. Given that the test distribution can dynamically change over time, OOD detection has become indispensable in high-stakes applications, such as healthcare and self-driving cars. Its ability to detect anomalous inputs and adapt to evolving scenarios makes it a vital component in ensuring the reliability and robustness of AI systems in real-world, dynamic environments.

\begin{figure}
    \centering
    \includegraphics[width=1.0\linewidth]{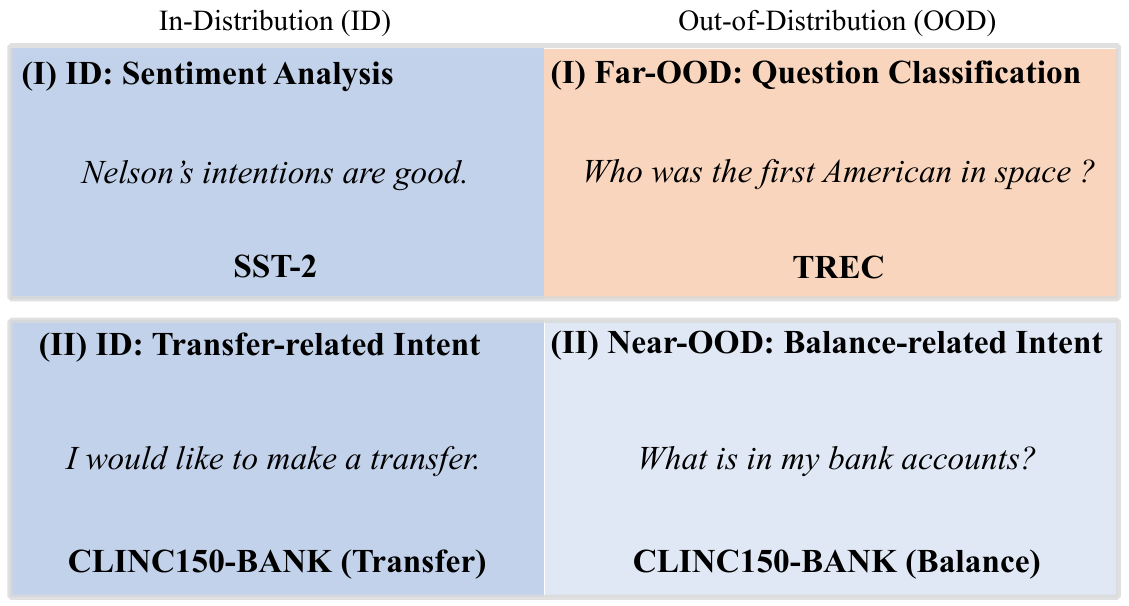}
    \caption{Illustration of two types of OOD instances compared to ID samples: far-OOD where ID and OOD data come from different domains and near-OOD where ID and OOD data come from the same domain but with different classes. }
    \label{fig:far_near_ood}
\end{figure}

Utilizing sentence representations yielded by pre-trained language models (PLMs) to derive OOD confidence scores has been the \emph{de facto} method for textual OOD detection. 
Specifically, PLMs are first fine-tuned on the ID data and then OOD detectors are applied on the sentence representations generated by PLMs. Compared to ID data, there are two types of OOD instances: far-OOD where ID and
OOD data come from different domains and near-OOD where ID and OOD data come from the same domain but with different classes, as shown in Figure~\ref{fig:far_near_ood}. Typically, near-OOD samples are harder to recognize.
A body of works~\cite{hendrycks2020pretrained,podolskiy2021revisiting, uppaal2023fine,zhan2024vi} have shown that Transformer-based models can produce better sentence representations for OOD detection.
However, these studies have mainly focused on evaluating the OOD detection performance of small-scale encoder-based Transformers, such as RoBERTa and BERT.

Recently, large language models (LLMs) have made significant strides in various cognitive tasks, yet their capabilities on OOD detection remain largely unexplored. Unlike relatively small-scale PLMs used by prior studies, LLMs often display notable differences.
In particular, the majority of previously prominent PLMs utilized for OOD detection adopt the encoder-based architecture, such as BERT and RoBERTa. These models are predominantly designed with a pre-training objective that focuses on sentence classification.
However, recent LLMs~\cite{touvron2023llama,zeng2022glm,du2022glm,chowdhery2022palm, chung2022scaling} exclusively adopt an autoregressive training objective during pre-training. 
Consequently, the hidden states of LLMs are specialized for next token prediction, which could influence their performance in OOD detection. 
Moreover, previous works test changes in OOD detection when adapting PLMs to downstream tasks through discriminative fine-tuning, even for decoder-based models~\cite{cho2023probing}. 
However, a more intuitive approach is to probe the pre-training knowledge of LLMs through generative fine-tuning, which better aligns LLMs' pre-training objective with downstream tasks.
Thus, it is imperative to extensively investigate the OOD detection capabilities of LLMs to gain deeper insights into their potential and limitations.

This paper aims to fill this gap by offering a comprehensive and structured assessment of OOD detection with LLMs across varying scales (ranging from 7B to 65B). 
Notably, our evaluation process is specifically designed to consider the scaling laws of LLMs with commonly utilized OOD detection detectors, ensuring broader and more generalized findings. In summary, our analysis has revealed the following new insights:

\paragraph{1. Discriminative \emph{vs.} generative fine-tuning.} We have observed that
generative fine-tuning demonstrates greater resilience to the issue of ID overfitting when compared to discriminative fine-tuning. As highlighted by \citet{uppaal2023fine}, there exists a trade-off between achieving higher accuracy on ID tasks and ensuring effective OOD detection. It has been shown that OOD detectors progressively lose efficacy as the training of ID tasks continues. However, our findings indicate that adopting a generative approach to fine-tuning LLMs can effectively mitigate this issue, potentially resulting in stable OOD performance even as training progresses and ID accuracy improves.

\paragraph{2. LLM-based far- \emph{vs.} near-OOD detection.} Our results consistently demonstrate that LLMs are natural far-OOD detectors. Remarkably, LLMs of all scales achieve near-perfect OOD performance in far-OOD scenarios without requiring any fine-tuning. 
However, when it comes to near-OOD detection, only the 65B model is able to achieve satisfactory performance without any fine-tuning.
Despite that, we discover that fine-tuning significantly improves the near-OOD detection capability of LLMs.

\paragraph{3. Anisotropy \emph{vs.} isotropy.} Our experimental results suggest that the cosine distance function, when used as a straightforward OOD detector, performs exceptionally well.
This observation leads to an intriguing discovery: the embedding spaces of LLMs exhibit a desirable isotropic property, which is not possessed by the BERT family models. The sentence embeddings produced by the BERT family models have been noted to possess an undesirable characteristic of being concentrated within a narrow cone, a phenomenon referred to as anisotropic representations~\cite{ethayarajh2019contextual}, which negatively affects tasks involving semantic relationships and is commonly known as representation degeneration~\cite{gao2019representation}. The issue is resolved through the isotropic representations generated by LLMs, which allow the cosine distance to excel in OOD detection and may potentially benefit a broad spectrum of tasks.

\begin{figure*}
    \centering
    \includegraphics[width=0.97\linewidth]{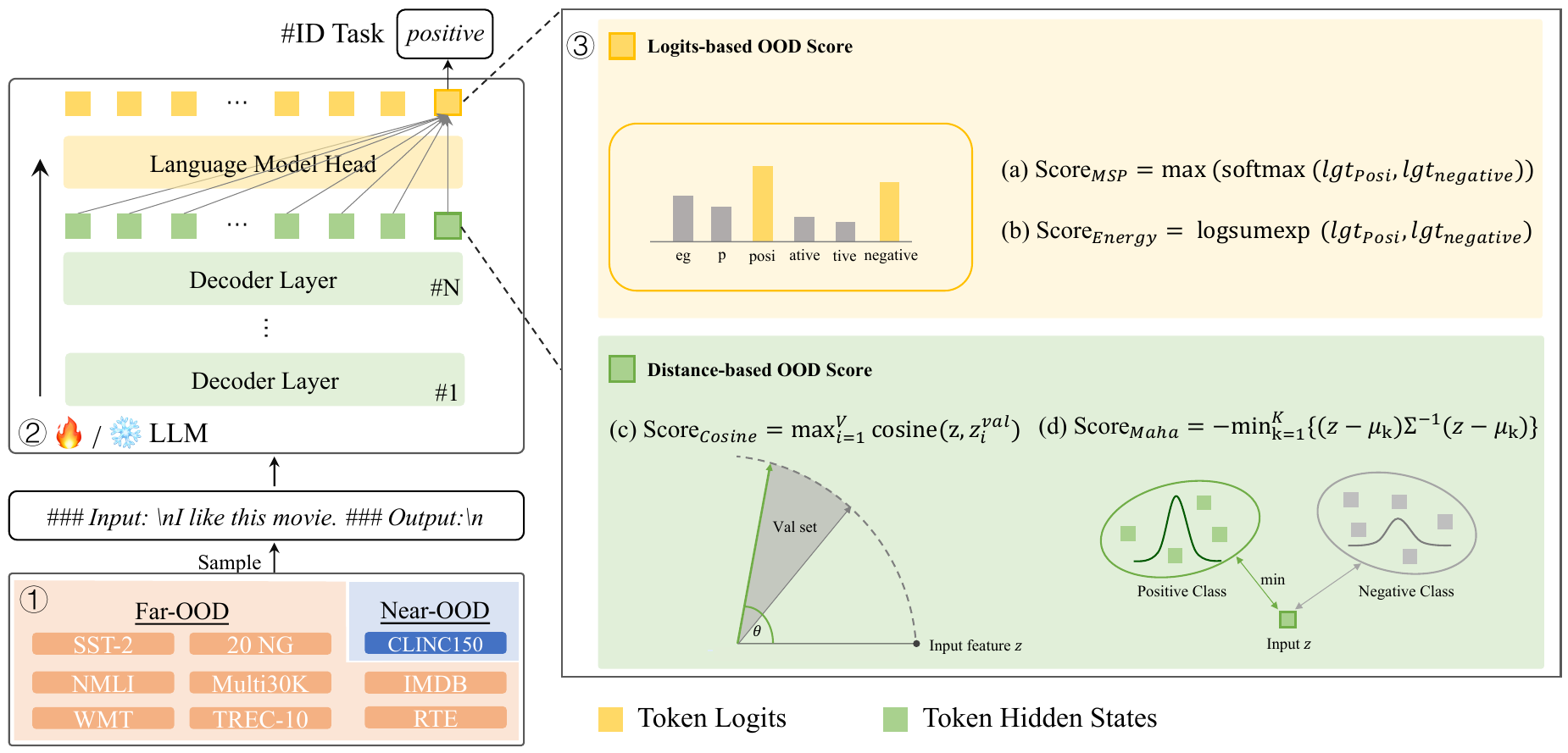}
    \caption{Our proposed evaluation framework for LLMs at OOD detection, taking three aspects into consideration: \ding{172} distribution of OOD samples (near or far), \ding{173} impact of model tuning on OOD detection, and \ding{174} diverse OOD score functions.}
    \label{fig:framework}
\end{figure*}

\section{Related Work}
\subsection{Out-of-Distribution Detection}
Out-of-Distribution (OOD) detection has a long history in machine learning and is highly related to research topics like outlier detection, anomaly detection and novelty detection~\cite{hendrycks2022x}. In the task setting of OOD detection, the \emph{in-distribution} is characterized by the labeled training dataset and the \emph{out-of-distribution} refers to anything else that possesses distributional shifts.
Note that OOD detection differs from outlier detection in  that it requires accurate classification of both of ID and OOD data~\cite{yang2021generalized}.

\subsection{Textual OOD Detection with PLMs}
The significance of textual OOD detection in ensuring the robustness of NLP applications, such as dialogue systems, has led to a surge in research interest. Pre-trained Transformers have shown intrinsic superiority in handing OOD detection~\cite{hendrycks2020pretrained, zhan2021out}. 

Several works have further evaluated the OOD performance of PLMs with respect to commonly used OOD detectors including MSP~\cite{DBLP:conf/iclr/HendrycksG17}, Mahalanobis distance (MD)~\cite{lee2018simple}, and Energy score~\cite{liu2020energy}. For example, \citet{podolskiy2021revisiting} show that the Gaussian distribution assumption of MD better matches the representation space of BERT and can yield the best OOD performance in intent OOD detection benchmarks. \citet{zhou2021contrastive} show that a contrastive regularizer can further improve the sentence representation of Transformers for OOD detection. 

More recently, \citet{uppaal2023fine} present a thorough analysis on the fine-tuning strategies for OOD detection with RoBERTa and show that RoBERTa~\cite{liu2019roberta} without fine-tuning can achieve near-perfect far-OOD detection performance. 
Similarly, we find that LLMs can also achieve perfect far-OOD detection performance without fine-tuning. 
\citet{cho2023probing} explore the OOD detection capability of medium-sized PLMs (such as GPT-2~\cite{radford2019language}), as well as the impact of various ID fine-tuning techniques. 
While they also assess decoder-based models, the models they evaluated are not extensive as this work and they neglect to undertake generative ID tuning, a crucial step to fine-tune decoder-based models for downstream ID tasks. Furthermore, the models they examine remain at relatively moderate scales, and an exploration of the possible data-efficient characteristics of the model is lacking.

Recently, large language models (LLMs) have been leading a paradigm shift in the field of natural language processing (NLP)~\cite{touvron2023llama,wei2022emergent,zeng2022glm,du2022glm,chowdhery2022palm, chung2022scaling, lu2021getting, feng2023towards}. The use of LLMs to solve NLP tasks in a generative way has become widespread.
These LLMs commonly adopt the decoder-based architecture and are trained with the autoregressive objective. In this paper, we focus on the OOD performance of open-source LLMs and anticipate our work can provide useful insights for OOD detection under this paradigm.

\section{Method}
\newcommand*\zero{\includegraphics[width=0.018\linewidth]{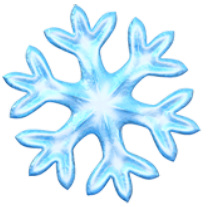}}
\newcommand*\tuned{\includegraphics[width=0.018\linewidth]{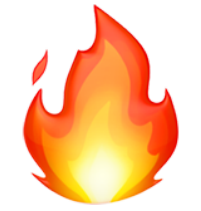}}
\newcommand*\zerol{\includegraphics[width=0.03\linewidth]{imgs/snow.pdf}}
\newcommand*\tunedl{\includegraphics[width=0.03\linewidth]{imgs/fire.pdf}}

\paragraph{Problem statement.} The objective of OOD detection is to effectively differentiate between instances that belong to a specific distribution (in-distribution $\mathcal{D}_{\text{ID}}$) and those falling outside of that distribution (out-of-distribution $\mathcal{D}_{\text{OOD}}$). To better and fairly evaluate the capabilities of LLMs for OOD detection compared to prior smaller models (e.g., RoBERTa~\cite{liu2019roberta})~\cite{uppaal2023fine}, we utilize the same sentence classification task as the ID training task. 
In practical application scenarios, undesired inputs (e.g., a severe distribution shift towards ID data) may occur, and an OOD confidence scoring function $f_\text{OOD}$ can be used to reject whether outputting results for such inputs or not.

\subsection{ID Generative Fine-tuning with LLMs}\label{sec:generative tuning}

For the ID sentence classification task, 
we align with the nature of LLMs and adopt a generative approach (referred to as open-ended classification)~\cite{radford2018improving}. Concretely, given an input sentence $\mathbf{X_s}$, we first expand it with a simple template: ``\text{\#\#\#} Input:\textbackslash n\textcolor{brown}{$\mathbf{X_s}$} \#\#\# Output:\textbackslash n'', to facilitate the extraction of outputs by identifying the section following the ``Output'' symbol. Subsequently, we maximize the probability of generating the target label $\mathbf{X_a}$ with $L$ tokens by:
\begin{equation} \label{eq:cls}
    \max p (\mathbf{X_a}|\mathbf{X_s}) = \prod_{i=1}^L \ p_\theta(x_i| \mathbf{X_s}, \mathbf{X}_{\mathbf{a}, <i}), 
\end{equation}
where $\theta$ represents the model parameters and $\mathbf{X}_{\mathbf{a}, <i}$ are partial label tokens that come before the current prediction token $x_i$.  


\paragraph{Parameter-efficient fine-tuning.} To improve the performance of LLMs in the in-distribution sentence classification task, we employ a parameter-efficient fine-tuning (PEFT) approach, to minimize the usage of additional parameters.
Specifically, we utilize the low-rank adaptation (LoRA)~\cite{hu2021lora} technique which freezes the pre-trained LLMs' weights and inserts trainable rank decomposition matrices into each Transformer layer. We perform PEFT with answer predictions, i.e.,  only class label tokens are utilized to compute the auto-regressive loss.
During the test stage, we use \emph{strict matching} to determine whether the generated labels are identical to the ground truth. 


\subsection{OOD Detection with LLMs}
The overview of our OOD detection framework is illustrated in Figure~\ref{fig:framework}. 
Our primary focus is on decoder-like LLMs, such as LLaMA~\cite{touvron2023llama}, as they have demonstrated excellent performance when their model size scales up~\cite{openai2023gpt4, brown2020language}.
To obtain a comprehensive observation, we conduct OOD detection experiments on two different semantic distribution settings~\cite{ming2022impact, lin2021mood}: far-OOD and near-OOD (\emph{cf.} Figure~\ref{fig:far_near_ood}). Regarding OOD detection methods, we focus on the prevailing post-hoc paradigm~\cite{yang2021generalized}. In the following, we elaborate on how to integrate post-hoc OOD detectors into decoder-style LLMs, which has not been addressed in existing literature.

\paragraph{Customized post-hoc methods.} According to prior studies~\cite{hendrycks2020pretrained, zhou2021contrastive}, there mainly exist two categories of post-hoc methods: logits-based OOD score functions and distance-based ones.
Since previous works used these methods for language models accompanied by a classifier, we here customize them for decoder-type LLMs with only a language model head (as shown in Figure~\ref{fig:framework}) in the following:

\textbf{Logits-based OOD score functions} operate on the final class-related logits. In generative classification, the generated class name is usually composed of several tokens, e.g., ``positive'' consists of ``posi'' and ``tive''. Instead of calculating the probability (logits) for the entire ID class name, we simplify the process by considering the probability assigned to the first token of its class name.   For instance, in a sentiment analysis task with classes like ``positive'' and ``negative'' as depicted in Figure~\ref{fig:framework}, we only need to identify the probability corresponding to the tokens ``posi'' and ``negative'' respectively. Considering that different class names may have common prefixes, such as ``positive'' and ``position'', we will rephrase the conflicting class names at the beginning of ID training, such as replacing ``position'' with ``location''. In practice, we observe this re-translation has no impact on the ID task. Overall, there are mainly two logits-based functions:
\begin{itemize}
    \item Maximum softmax probability (MSP)~\cite{lee2018simple} utilizes the maximum softmax probability corresponding to each class, i.e, score $\mathcal{S}(x) = \max \{p(y_i|x)\}_i^K$, where $K$ is the number of classes, and ID samples always exhibit higher probability scores while OOD ones correspond to lower scores. 

    \item Energy score (Energy)~\cite{liu2020energy,lecun2006tutorial} computes confidence score
    $\mathcal{S}(x) = \log \sum_{i}^{K} e^{(w^{T} \cdot z)_i}$
    where $w^{T}$ is the weight of the language model head and $z$ is all word embeddings. Note that for both MSP and Energy, we only select the probability and logits corresponding to the first token of each class name, as mentioned above.
    
\end{itemize}


\noindent
\paragraph{Distance-based OOD score functions} apply to sentence representations. Prior studies using encoder-based PLMs treated the embeddings of special token <cls> as sentence representations. 
For LLMs, we employ the embeddings of the last token as the representation. There are mainly two functions considered for evaluation: Mahalanobis distance (Maha)~\cite{lee2018simple} and Cosine distance (Cosine)~\cite{zhou2021contrastive}\footnote{We refer authors to original papers for more details.}.

\begin{table*}[t]
\centering
\resizebox{\linewidth}{!}{
\begin{tabular}{ccccccccccccccc}
\hline
\hline

& &  &  \multicolumn{3}{c}{\bf Maha} & \multicolumn{3}{c}{\bf Cosine} & \multicolumn{3}{c}{\bf MSP} & \multicolumn{3}{c}{\bf Energy}\\
 \cmidrule(lr){4-6} \cmidrule(lr){7-9} \cmidrule(lr){10-12} \cmidrule(lr){13-15}
& ID Dataset & LLM & AUROC  $\uparrow$  & FAR@95  $\downarrow$& AUPR $\uparrow$ &  AUROC  $\uparrow$  & FAR@95 $\downarrow$ & AUPR $\uparrow$ & AUROC  $\uparrow$  & FAR@95  $\downarrow$& AUPR $\uparrow$ &  AUROC  $\uparrow$  & FAR@95  $\downarrow$& AUPR $\uparrow$ \\
\specialrule{0.05em}{0.3em}{0.3em}

\multirow{8}{*}{\rotatebox{90}{Far-OOD}} & \multirow{4}{*}{SST-2} 
& LLaMA-7B & 0.991 & 0 & 0.993 & 0.990 & 0.006 & 0.990 & 0.905 & 0.318 & 0.811 & 0.368 & 0.930 & 0.380\\
 & &  LLaMA-13B & 0.992 & 0 & 0.993 & 0.990 & 0.005 & 0.989 & 0.939 & 0.213 & 0.818 & 0.571 & 0.778 & 0.478\\
& &  LLaMA-30B & 0.994 & 0.003 & 0.993 & 0.991& 0.009 & 0.990 & 0.881 & 0.361 & 0.742 & 0.651 & 0.738 & 0.540\\
& &  LLaMA-65B & 0.991 & 0.007 & 0.991 & 0.990 & 0.007 & 0.992 & 0.776 & 0.621 & 0.646 & 0.544 & 0.821 & 0.485\\

\cmidrule(lr){2-15}

&  \multirow{4}{*}{20NG} 
& LLaMA-7B & 0.997 & 0 & 0.995 & 0.998 & 0 & 0.996  & 0.441 & 0.929 & 0.391 &0.571 & 0.784 & 0.417 \\
& &  LLaMA-13B & 0.996 & 0.006 & 0.989 & 0.993 & 0.004 & 0.990 & 0.622 & 0.754 & 0.482 & 0.491 & 0.932 & 0.362 \\
& &  LLaMA-30B & 0.995 & 0.005 & 0.987 & 0.995 & 0.002 & 0.993 & 0.533 & 0.847 & 0.424 & 0.491 & 0.906 & 0.362 \\
& &  LLaMA-65B & 1 & 0 & 0.998 & 0.999 & 0 & 0.997 & 0.616 & 0.764 & 0.421 & 0.508 & 0.925 & 0.369\\

\specialrule{0.05em}{0.3em}{0.3em}

\multirow{8}{*}{\rotatebox{90}{Near-OOD}} & \multirow{4}{*}{CLINC-Banking} 
& LLaMA-7B & 0.896 & 0.568 & 0.921 & 0.891 & 0.587 & 0.916 &  0.720 & 0.814 & 0.763 & 0.722 & 0.818 & 0.758\\
& &  LLaMA-13B & 0.905 & 0.408 & 0.922 & 0.903 & 0.514 & 0.922 & 0.739 & 0.769 & 0.760 & 0.713 & 0.831 & 0.743\\
& &  LLaMA-30B & 0.895 & 0.472 & 0.913 & 0.910 & 0.424 & 0.923 & 0.733 & 0.813 & 0.746 & 0.724 & 0.795 & 0.735\\
& &  LLaMA-65B & 0.951 & 0.255 & 0.964 & 0.956 & 0.200 & 0.964 & 0.823 & 0.604 & 0.834 & 0.826 & 0.614 & 0.834\\
\cmidrule(lr){2 - 15}
& \multirow{4}{*}{CLINC-Travel} 
& LLaMA-7B & 0.895 & 0.680 & 0.932 & 0.887 & 0.738 & 0.927 & 0.584 & 0.921 & 0.640 & 0.637 & 0.912 & 0.674\\
& &  LLaMA-13B & 0.942 & 0.485 & 0.964 & 0.922 & 0.730 &  0.955 & 0.639 & 0.834 & 0.696 & 0.633 & 0.909 & 0.695\\
& &  LLaMA-30B & 0.926 & 0.458 & 0.950 & 0.928 & 0.523 & 0.950 & 0.650 & 0.911 & 0.697 & 0.653 & 0.888 & 0.698\\
& &  LLaMA-65B & 0.959 & 0.182 & 0.971 & 0.976 & 0.076 & 0.986 & 0.739 & 0.745 & 0.753 & 0.755 & 0.681 & 0.768\\

\hline
\hline
\end{tabular}
}
\caption{
 OOD detection performance of zero-grad LLaMA models. We use the full validation set to calculate each OOD score. The results are averaged over five seeds.  
}
\label{tab:zero_grad_llama}
\end{table*}

\begin{figure*}
    \centering
    \includegraphics[width=0.97\linewidth]{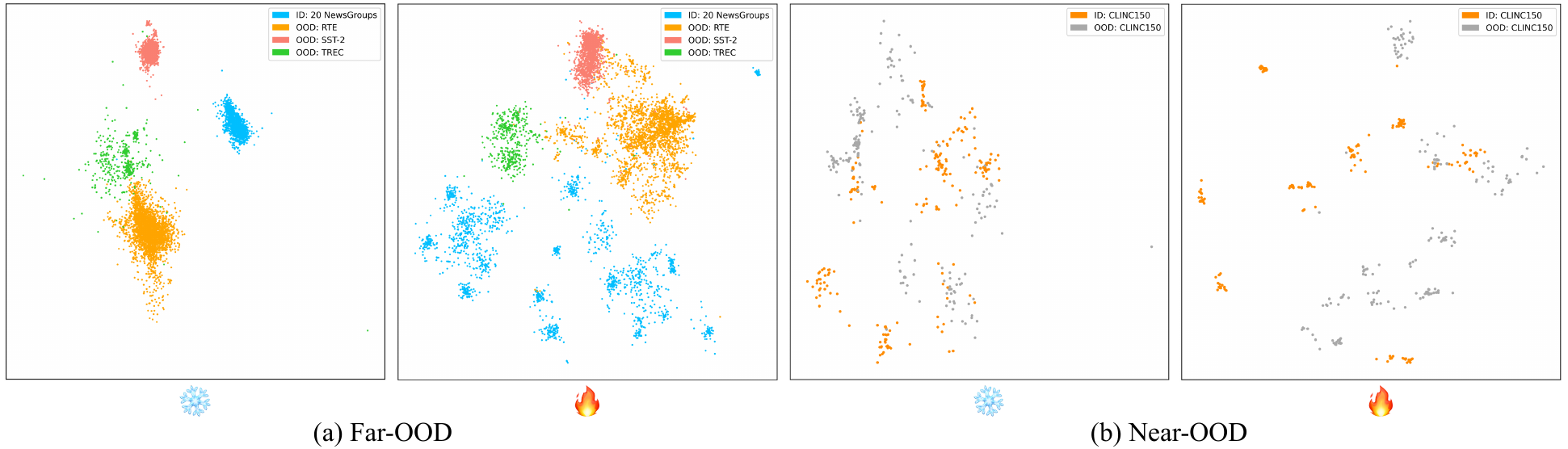}
    \caption{UMAP~\cite{mcinnes2018umap} visualization of representations generated from the penultimate layer of the zero-grad (\zero{}) and fine-tuned (\tuned{}) LLaMA-7B models. (a) Far-OOD: 20NG is treated as ID while SST-2, RTE, and TREC are treated as OOD. (b) Near-OOD: the banking domain of CLINC150 is selected, of which 50\% of the classes are treated as ID, and the rest are treated as OOD.
    }
    \label{fig:zero_fine_embedding}
\end{figure*}

\section{Experimental Setup}\label{sec:exp}
\subsection{Datasets}
To draw universal conclusions, we conduct a comprehensive evaluation of two kinds of dataset distribution settings~\cite{arora2021types} as illustrated in Figure~\ref{fig:far_near_ood} and Figure~\ref{fig:framework}. 
\paragraph{Far-OOD.} In this paradigm, ID and OOD samples come from different distributions (datasets), exhibiting significant semantic differences. Following \citet{hendrycks2020pretrained} and \citet{zhou2021contrastive}, we evaluate 8 datasets, including \texttt{20 Newsgroups (20NG)}~\cite{lang1995newsweeder} for topic classification, \texttt{RTE}~\cite{wang2018glue} and \texttt{MNLI}~\cite{williams2017broad} for nature language inference, \texttt{TREC-10}~\cite{li2002learning} for question classification, \texttt{SST-2}~\cite{socher2013recursive} and \texttt{IMDB}~\cite{maas2011learning} for sentiment analysis, and the English side of \texttt{Multi30K}~\cite{elliott2016multi30k} and \texttt{WMT16}~\cite{bojar-etal-2016-findings} for machine translation. Among them, we choose 20NG and SST-2 as two separate \emph{in-distribution} tasks and the remaining ones are recognized as \emph{out-distribution}. 
Note that when SST-2 is used as the ID, we do not consider IMDB as an OOD dataset since both of them are sentiment analysis tasks.

\paragraph{Near-OOD.} We also test on a more challenging scenario, where ID and OOD samples come from the same domain but with disjoint label sets. A well-researched domain is OOD intent detection~\cite{larson2019evaluation}. Specifically, we use \texttt{CLINC150} dataset and choose \emph{Banking} and \emph{Travel} domains. Within each domain, 50\% of the classes are chosen as ID, and the remaining classes as OOD. 

\subsection{Evaluation Metrics} We employ three commonly used metrics for OOD detection: (1) AUROC (area under the receiver operating characteristic curve). (2) FAR@95 (false alarm rate at 95\% recall). It represents the probability of incorrectly classifying a negative sample as positive when the Recall or True Positive Rate (TPR) is 95\%. We treat the OOD class as negative. (3) AUPR (area under the precision-recall curve). Additionally, we use accuracy as a metric for ID classification task.
\begin{table}[h]
\centering
\resizebox{\linewidth}{!}{
\begin{tabular}{c|cccc}
\hline
Dataset& Full-shot & 10-shot & 5-shot & 1-shot\\
\hline
SST-2  & 16 & 8 & 4 & 2\\
20NG & 8 & 8 & 8 & 8\\
\makecell{CLINC150 \\ (Banking or Travel)}  & 16 &16 & 16 & 8\\

\hline
\end{tabular}
}
\caption{
Batch size configuration for each dataset.
}
\label{tab:bsz}
\end{table}

\begin{table*}[t]
\centering
\resizebox{\linewidth}{!}{
\begin{tabular}{cccccccccccccccc}
\hline
\hline

& &  & & \multicolumn{3}{c}{\bf Maha} & \multicolumn{3}{c}{\bf Cosine} & \multicolumn{3}{c}{\bf MSP} & \multicolumn{3}{c}{\bf Energy}\\
 \cmidrule(lr){5-7} \cmidrule(lr){8-10} \cmidrule(lr){11-13} \cmidrule(lr){14-16}
& ID Dataset & Shot & ID ACC$ \uparrow$ & AUROC  $\uparrow$  & FAR@95  $\downarrow$& AUPR $\uparrow$ &  AUROC  $\uparrow$  & FAR@95 $\downarrow$ & AUPR $\uparrow$ & AUROC  $\uparrow$  & FAR@95  $\downarrow$& AUPR $\uparrow$ &  AUROC  $\uparrow$  & FAR@95  $\downarrow$& AUPR $\uparrow$ \\
\specialrule{0.05em}{0.3em}{0.3em}

\multirow{8}{*}{\rotatebox{90}{Far-OOD}} & \multirow{4}{*}{SST-2} 
& 1 & 0.535 & 0.5 & 1.0 & 0.422 & 0.954 & 0.250 & 0.934 & 0.664 & 0.581 & 0.587 & 0.716 & 0.589 & 0.637  \\
& &  5 & 0.664 & 0.878 & 0.625 & 0.843 & 0.973 & 0.045 & 0.971 & 0.768 & 0.493 & 0.674 & 0.885 & 0.408 & 0.794\\
& &  10 & 0.857 & 0.967 & 0.204 & 0.962 & 0.991 & 0.009 & 0.987 & 0.771 & 0.514 & 0.693 & 0.896 & 0.379 & 0.803\\
& &  Full & 0.976 & 0.993 & 0.004 & 0.992 & 0.993 & 0.005 & 0.991 & 0.947 & 0.298 & 0.888 & 0.961 & 0.189 & 0.907\\

\cmidrule(lr){2-16}

& \multirow{4}{*}{20NG} 
& 1 & 0.463 & 0.5 & 1 & 0.380 & 0.991 & 0.047 &  0.985 & 0.756 & 0.779 & 0.670 & 0.850 & 0.681 & 0.824\\
& &  5 & 0.713 & 0.983 & 0.074 & 0.975 & 0.991 & 0.023 & 0.989 & 0.868 & 0.503 & 0.799 & 0.947 & 0.283 & 0.918\\
& &  10 & 0.796 & 0.992 & 0.042 & 0.987 & 0.996 & 0.013 & 0.991 & 0.893 & 0.438 & 0.840 & 0.951 & 0.215 & 0.924 \\
& &  Full & 0.944 & 0.995 & 0.003 & 0.991 & 0.993 & 0.007 & 0.991 & 0.959 & 0.207 & 0.939 & 0.968 & 0.114 & 0.945\\

\specialrule{0.05em}{0.3em}{0.3em}

\multirow{8}{*}{\rotatebox{90}{Near-OOD}}  & \multirow{4}{*}{CLINC-Banking} 
& 1 & 0.589 & 0.5 & 1 & 0.533 & 0.905 & 0.510 & 0.926 & 0.846 & 0.696 & 0.860 & 0.870 & 0.658 & 0.897\\
& &  5 & 0.882 & 0.863 & 0.614 & 0.879 & 0.962 & 0.255 & 0.968 & 0.873 & 0.556 & 0.878 & 0.903 & 0.463 & 0.916\\
& &  10 &0.949 & 0.937 & 0.424 & 0.956 & 0.968 & 0.157 & 0.974 & 0.902 & 0.422 & 0.902 & 0.919 & 0.346 & 0.929\\
& &  Full & 0.973 & 0.958 & 0.231 & 0.969 & 0.964 & 0.147 &0.970 &0.936 & 0.269 &0.945 & 0.930 & 0.225 & 0.931\\

\cmidrule(lr){2-16}

& \multirow{4}{*}{CLINC-Travel}
& 1 & 0.526 & 0.5 &  1 &0.533 & 0.910 & 0.481 & 0.925 & 0.767 & 0.756 & 0.771 & 0.780 & 0.733 & 0.793\\
& &  5 & 0.964 & 0.897 & 0.644 & 0.925 & 0.974 & 0.148 & 0.983 & 0.886 & 0.415 & 0.886 & 0.875 & 0.420 & 0.872\\
& &  10 & 0.984 & 0.975 & 0.137 & 0.983 & 0.982 & 0.078 & 0.988 & 0.930 & 0.3 & 0.931 & 0.933 & 0.231 & 0.933\\
& &  Full & 0.991 & 0.980 &  0.045 & 0.988 & 0.978 & 0.049 & 0.987 & 0.942 & 0.121 & 0.933 & 0.948 & 0.112 & 0.953\\

\hline
\hline
\end{tabular}
}
\caption{
The performance of the fine-tuned LLaMA-7B model for OOD detection and ID classification. ``Shot'' denotes the number of examples in the ID training or validation set.  We report the average results of five seeds. 
}
\label{tab:fine_tune_llama}
\end{table*}

\subsection{Implementation Details} \label{sec:implementation}
All experiments are conducted on a workstation with 4 NVIDIA A100 80G GPUs.
For zero-grad OOD detection, LLaMA-7B, -13B, -30B, and -65B are deployed on 1, 1, 2, and 4 A100 GPUs, respectively. 
When further fine-tuning LLMs on ID tasks, the LoRA configurations (Section~\ref{sec:generative tuning}) are that rank $r$ is 16, scaling $\alpha$ is 16, and query/key/value/output projection matrices $\{ W_q, W_k, W_v, W_o\}$ in each self-attention module need to be updated.
We train the network for 50 epochs with early stop criteria that if the model's performance on the validation set continuously drops for 6 consecutive epochs and the current epoch number exceeds 15, training will be terminated. 
We use AdamW optimizer with learning rate $1\times10^{-4}$, further decayed by linear schedule.  Due to the varying lengths of sentences in different ID datasets, we configure different batch sizes shown in Table~\ref{tab:bsz}. All experiments are conducted over five seeds (1, 2, 3, 4, 5).

\section{Findings}

\subsection{Zero-grad OOD Detection with LLMs}\label{sec:zero_grad}
In this section, we evaluate the \emph{zero-grad} OOD performance of LLMs.
The objective is to examine how well OOD detection performs when utilizing the knowledge acquired by LLMs during pre-training.
The results are summarized in Table~\ref{tab:zero_grad_llama} and all LLMs are frozen in this setting. Note that we use full-shot validation set to calculate each OOD score.

\begin{figure}[t]
    \centering
    \includegraphics[width=1\linewidth]{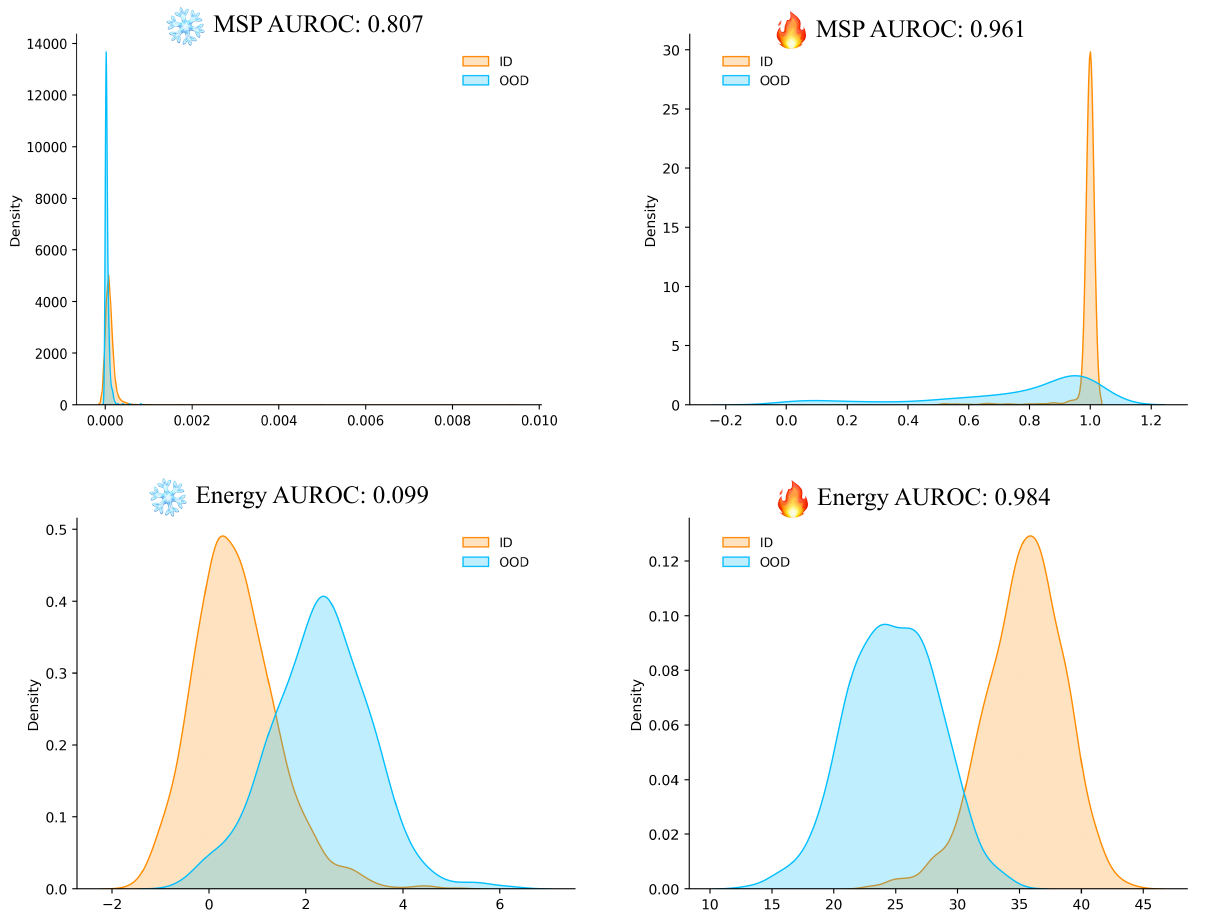}
    \caption{Impact of fine-tuning on logits-based OOD scores (MSP at the top row and Energy at the bottom row). We plot SST-2 (ID) \emph{vs.} TREC-10 (OOD) for visualization. }
    \label{fig:density}
\end{figure}

\paragraph{LLMs are natural far-OOD detectors.} As shown in Table~\ref{tab:zero_grad_llama}, when applying distance-based OOD detection methods, such as Maha and Cosine, all LLMs can achieve near-perfect results (e.g., AUROC and AUPR approach 1 while FAR@95 approaches 0). To better understand why distance-based OOD detectors are so effective, we visualize the corresponding sentence representations yielded by the penultimate layer (before the top head layer), as shown in Figure~\ref{fig:zero_fine_embedding} (a~\zerol{}). It can be found that representations from the same dataset are tighter, while ID and OOD sentences have clear boundaries, indicating the profound semantic discrimination prowess exhibited by LLMs. 
However, both MSP and Energy generate poor results. This is foreseeable, as both of them condition on the first token generated from the input sentence. When the model has not been fine-tuned, it often struggles to accurately output class names, leading to inferior OOD performance. Moreover, from the probability density of Figure~\ref{fig:density} (\zerol{}), it can be found that there is a significant overlap between ID and OOD, leading to a decrease in OOD detection performance.

\paragraph{The capability of LLMs for near-OOD detection improves with their scale.} 
We present the zero-grad near-OOD results in Table~\ref{tab:zero_grad_llama} (CLINC-Banking and CLINC-Travel).
For the near-OOD setting, as the number of model parameters increases, the OOD detection performance will also be improved. 
Remarkably, when the model has an exceedingly large number of parameters (i.e., LLaMA-65B), we can observe a dramatic performance surge~\cite{wei2022emergent} to detect OOD inputs, especially with distance-based OOD methods. In particular, the AUROC values for Maha and Cosine both surpass 95\%, and FAR95 is enhanced by \textbf{at least 30\%} in comparison to the 7B model. 

Furthermore, it is evident that the near-OOD performance of LLMs is notably inferior compared to their performance on far-OOD instances. To understand this,
we provide a visualization for this setting as illustrated in Figure~\ref{fig:zero_fine_embedding} (b~\zerol{}). The embeddings of ID and OOD samples are mixed up since their labels come from the same domain (i.e., travel or banking). Consequently, detecting near-OOD instances becomes notably more challenging than far-OOD instances.

\begin{figure*}
    \centering
    \includegraphics[width=1\linewidth]{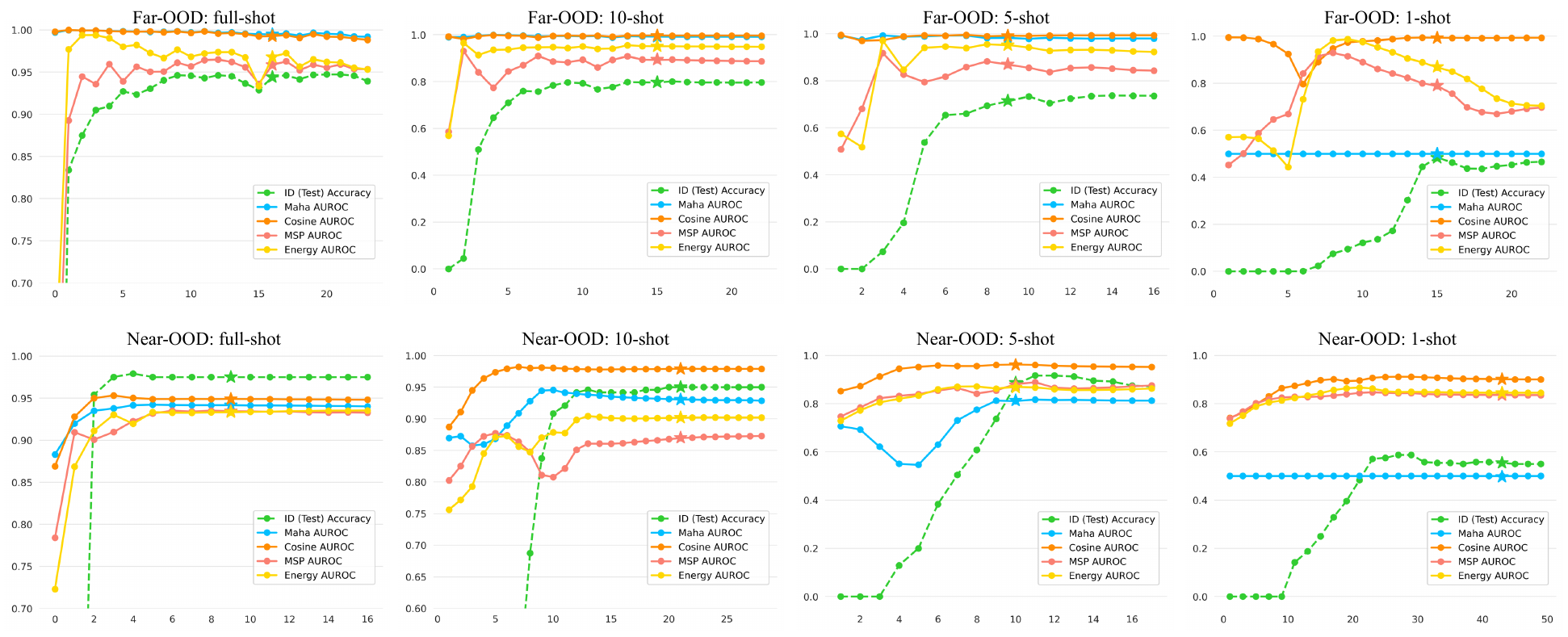}
    \caption{Performance changes for ID classification and OOD detection as training progresses with the different number of training shots. Top row: 20NG is ID training task; Bottom row: banking domain of CLINC150 is selected where 50\% classes are used as ID training task and the rest are OOD samples. The star ($\star$) on each line indicates the selected results whose epoch corresponds to the best ID performance on the validation set.}
    \label{fig:training_process}
\end{figure*}

\subsection{OOD Detection with Generatively Fine-tuned LLMs }\label{sec:analysis_generative}
In this subsection, we study the influence of fine-tuning LLMs on OOD detection. Specifically, we conduct an in-depth examination of how the OOD detection performance evolves with the progression of ID task training.

\paragraph{ID fine-tuning can boost OOD detection.} 
We fine-tune LLMs in a generative manner in both few-shot and full-shot scenarios. The results are summarized in Table~\ref{tab:fine_tune_llama}. Likewise, we present both far- and near-OOD results comparable to the zero-grad configuration. Clearly, fine-tuning LLMs on in-distribution tasks can notably augment the models' capacity to detect OOD instances, surpassing the performance of the zero-grad setting by a significant margin in most cases like in near-OOD setting and with logits-based functions (in both full-shot scenarios with LLaMA-7B model). 

In Figure~\ref{fig:training_process}, we present the fine-tuning curves. It can be observed that as the ID accuracy increases, almost all OOD detectors are improved accordingly. To study how fine-tuning impacts the ID \emph{vs.} OOD separability, we plot their density distributions in Figure~\ref{fig:density}. Clearly, fine-tuning can improve the separability between ID and OOD instances. A similar effect can be cross-validated in Figure~\ref{fig:zero_fine_embedding} (b~\tunedl{}) in which the embedding of different classes within the ID becomes more compact, while the separation between ID and OOD becomes clearer.
However, it is important to highlight that as the training continues, there is a possibility of encountering overfitting, which could result in inferior OOD performance, especially for logits-based methods as illustrated in Figure~\ref{fig:training_process} for both full-shot and 1-shot far-OOD scenarios. This observation is similar to the findings in \cite{uppaal2023fine}. 

\begin{figure}
    \centering
    \includegraphics[width=1\linewidth]{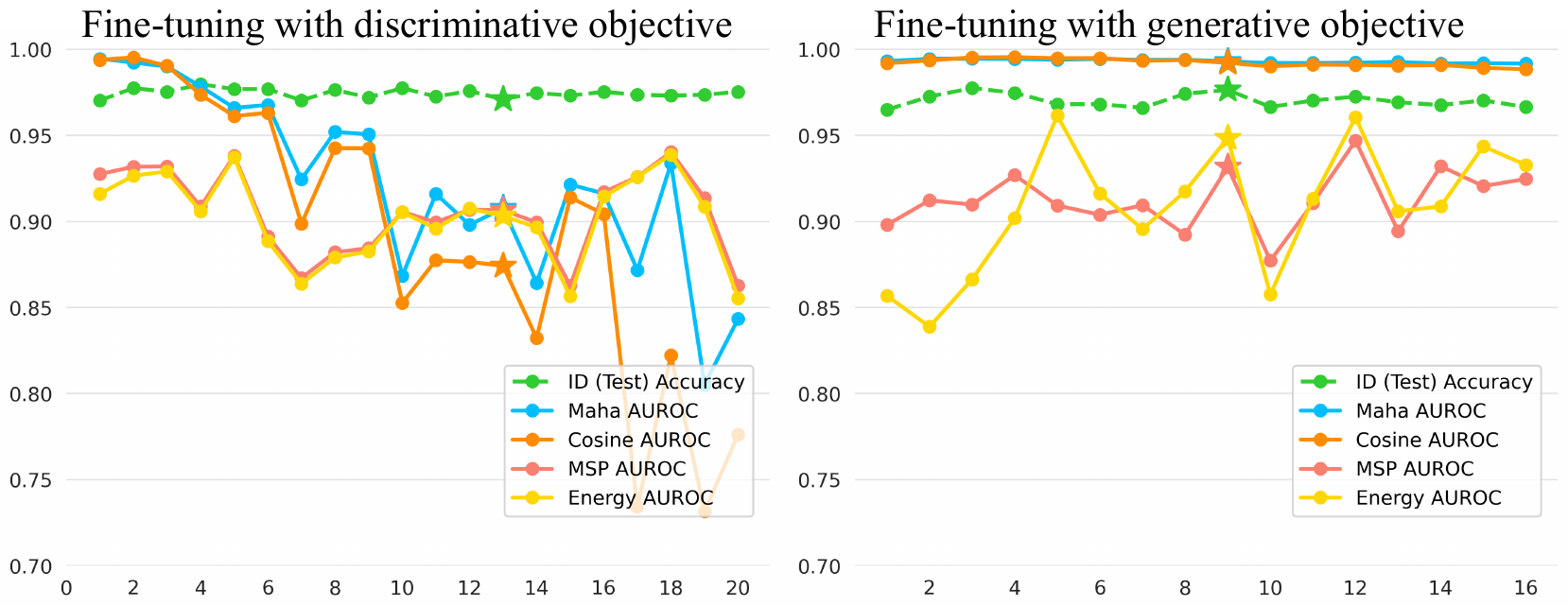}
    \caption{Impact of different ID training objectives, discriminative \emph{vs.} generative. SST-2 dataset with full data is used as the ID training task. }
    \label{fig:dis_and_gen}
\end{figure}

\begin{table*}[t]
\centering
\resizebox{0.9\linewidth}{!}{
\begin{tabular}{ccccccccccc}
\hline
\hline

&  &  \multicolumn{2}{c}{\bf Maha} & \multicolumn{2}{c}{\bf Cosine} & \multicolumn{2}{c}{\bf MSP} & \multicolumn{2}{c}{\bf Energy}\\
 \cmidrule(lr){3-4} \cmidrule(lr){5-6} \cmidrule(lr){7-8} \cmidrule(lr){9-10}
ID Dataset & PTM & AUROC  $\uparrow$  & FAR@95  $\downarrow$&  AUROC  $\uparrow$  & FAR@95 $\downarrow$  & AUROC  $\uparrow$  & FAR@95  $\downarrow$& AUROC  $\uparrow$  & FAR@95  $\downarrow$\\
\specialrule{0.05em}{0.3em}{0.3em}
& & \multicolumn{8}{c}{\bf \emph{Zero-grad}}\\
 
\multirow{4}{*}{SST-2}  & RoBERTa-L$\dagger$~\cite{uppaal2023fine} & 0.971 & 0.152 &0.919 & 0.414& - & - & - & - \\

& LLaMA-7B & \textbf{0.991} & \textbf{0} & \textbf{0.990} & \textbf{0.006} & 0.905 & 0.318 & 0.368 & 0.930 \\

\cmidrule(lr){2 - 10}
& & \multicolumn{8}{c}{\bf \emph{Fine-tuned}}\\

 & RoBERTa-L$\ddagger$~\cite{zhou2021contrastive} & 0.969 & 0.183 & 0.962 & 0.236 & 0.889 & 0.613 & 0.877 & 0.632\\
& LLaMA-7B & \textbf{0.993} & \textbf{0.004} & \textbf{0.993} & \textbf{0.005} & \textbf{0.947} & \textbf{0.298} & \textbf{0.961} & \textbf{0.189}\\

\specialrule{0.05em}{0.3em}{0.3em}

& & \multicolumn{8}{c}{\bf \emph{Zero-grad}}\\
 \multirow{4}{*}{20NG}  & RoBERTa-L$\dagger$~\cite{uppaal2023fine} & \textbf{0.998} & 0.002 & 0.998 & 0.002& - & - & - & - \\ 
& LLaMA-7B & 0.997 & \textbf{0} & \textbf{0.998} & \textbf{0} & 0.441 & 0.929 & 0.571 & 0.784\\

\cmidrule(lr){2 - 10}
  & & \multicolumn{8}{c}{\bf \emph{Fine-tuned}}\\

 & RoBERTa-L$\ddagger$~\cite{zhou2021contrastive} & 0.983 & 0.073 & 0.978 & 0.107 & 0.946 & 0.305 & 0.965 & 0.158\\
& LLaMA-7B & \textbf{0.995} & \textbf{0.003} & \textbf{0.993} & \textbf{0.007} & \textbf{0.959} & \textbf{0.207} & \textbf{0.968} & \textbf{0.114}\\
\hline
\hline
\end{tabular}
}
\caption{
Comparison of large and small PTMs under zero-grad and fine-tuned settings for OOD detection. $\dagger$ denotes the results we reproduce due to different calculating methods, while $\ddagger$ indicates results cited from the original paper.
}
\label{tab:sota}
\end{table*}

\paragraph{Generative fine-tuning generalizes better.} In addition to generative fine-tuning, we also explore discriminative fine-tuning by appending a classifier after LLMs (replacing the language model head\footnote{We use \texttt{LlamaForSequenceClassification} provided by Huggingface~\cite{wolf2019huggingface}}) to conduct ID task. The comparison of the trend charts presented in Figure~\ref{fig:dis_and_gen} reveals that generative fine-tuning tends to be less overfit on the ID task and all OOD detectors consistently perform better than discriminative fine-tuning, especially for distance-based OOD detectors.
To better understand this effect, based on the transformations of embeddings illustrated in Figure~\ref{fig:zero_fine_embedding} (a), it becomes evident that throughout the generative training process, while the ID's distribution shifts into class-specific clusters, a distinct separation continues to exist between these clusters and the OOD samples. This preserves the effectiveness of distance-based OOD detection methods. 
Prior study~\cite{uppaal2023fine} pointed out that discriminative tuning the small models (e.g., RoBERTa~\cite{liu2019roberta}) negatively impacts the performance of distance-based OOD detection methods. This issue also exists in discriminative tuning LLMs but has been solved in the generative tuning.

Besides, in Table~\ref{tab:sota}, we compare encoder-based and decoder-based Transformers and observe impressive improvement on decoder-based LLMs.

\begin{table}[t]
\centering
\resizebox{\linewidth}{!}{
\begin{tabular}{c|ccc}
\hline
\hline
& \multicolumn{3}{c}{Data Corpus} \\
PTMs &  CLINC150 & CLINC150-Banking & CLINC150-Travel \\
\hline
LLaMA-7B &  \textbf{0.4731} & \textbf{0.5529} & \textbf{0.5312}\\
RoBERTa-L & 0.9991 & 0.9992 & 0.9989 \\
\hline
\hline
\end{tabular}
}
\caption{Average sentence anisotropy of model's last layer.}
\label{tab:anisotropy}
\end{table}

\paragraph{Cosine distance is a data-efficient OOD detector.} To further investigate whether LLMs possess data-efficient OOD detection capabilities, we configure the training samples of the ID as few-shot instances (e.g., 1, 5, and 10). Please note that we also set the number of validation sets to be the same shot, since all OOD detection methods rely on the validation set. Results presented in Table~\ref{tab:fine_tune_llama} convey that as the number of shots increases, the OOD detection capability of the LLMs also improves. Moreover, distance-based OOD detection methods are superior to logits-based ones, and they can achieve good performance even with only 10-shot samples. Particularly, cosine distance is a data-efficient OOD detector that can provide effective detection by requiring only \textbf{1-shot} instance. For example, it achieves AUROC of \textbf{99.1\%} (near-perfect) on 20NG (ID) and over 90\% on others. Besides, in the 1-shot setting, the Mahalanobis distance loses its efficacy since it's unfeasible to model the necessary Gaussian distribution when there's only a single sample for each class.

\paragraph{Isotropy \emph{vs.} anisotropy.} By examining Table~\ref{tab:zero_grad_llama} and Table~\ref{tab:fine_tune_llama}, it becomes evident that Cosine distance, as a simple OOD detector, consistently delivers superior performance and ranks among the top performers in both the zero-grad and generative fine-tuning settings. We provide an explanation of this phenomenon from the perspective of representation learning. In the past few years, the anisotropic issue, also known as the representation degeneration problem, of BERT family models has garnered considerable attention~\cite{ethayarajh2019contextual,gao2019representation}. Researchers have highlighted that BERT's sentence representations are concentrated within a narrow cone, resulting in substantial challenges for tasks involving semantic matching.
Nevertheless, we discover that this concern does not apply to LLMs. The representations generated by off-the-shelf LLMs inherently exhibit isotropy, enabling Cosine distance to excel in OOD detection.
To quantify anisotropy, we adopt the methodology introduced by \citet{ethayarajh2019contextual} to measure sentence-level anisotropy. Let $\mathbf{X_i}$ be a sentence that appears in the corpus. The anisotropy value can be calculated by:
    \begin{flalign}
    \begin{aligned}\label{eq:isotropy}
    \text{anisotropy} = \frac{1}{n^2-n}\left|\sum_i \sum_{j\neq i} \cos(z(\mathbf{X_i}), z(\mathbf{X_j})) \right|, 
    \end{aligned}
    \end{flalign}
where $\cos$ is the cosine similarity and $z(\cdot)$ is the sentence embedding from the last layer. A higher anisotropy value suggests that the sentence embeddings are less distinguishable by Cosine distance.
The quantitative results presented in Table~\ref{tab:anisotropy} show that the anisotropy values of LLMs are considerably lower in comparison to those of RoBERTa.

\section{Analysis}
\subsection{Performance of More LLMs}

\begin{table}[h]
\centering
\resizebox{\linewidth}{!}{
\begin{tabular}{cccccc}
\hline
\hline
ID Dataset & PTMs &  Maha & Cosine & MSP & Energy \\
\hline
& & \multicolumn{4}{c}{\bf \emph{Zero-grad}}\\
\multirow{3}{*}{SST-2} & OPT-6.7B &  0.982 &  0.983  &  0.413  &  \textbf{0.571} \\
& LLaMA-7B & \textbf{0.991} &  0.990  &  0.905  &  0.368  \\
& LLaMA2-7B &  \textbf{0.991} &  \textbf{0.997} & \textbf{0.917}  &  0.516 \\

\hline

& & \multicolumn{4}{c}{\bf \emph{Fine-tuned}}\\
\multirow{3}{*}{\shortstack{CLINC-Banking \\ (Full)}} & 
OPT-6.7B & 0.921 &  0.932  &  0.915  &  0.922 \\
&  LLaMA-7B & 0.958 &  0.964  &  0.936  &  0.930 \\
&  LLaMA2-7B & \textbf{0.967} &  \textbf{0.970}  &  \textbf{0.944}  &  \textbf{0.937} \\
\hline
\hline
\end{tabular}
}
\caption{OOD detection performance of various LLMs. AUROC result is reported.}
\label{tab:various-llms}
\end{table}

Additionally, we test OPT-6.7B~\cite{zhang2022opt} and LLaMA2-7B~\cite{touvron2023llama2} here with settings that zero-grad OOD performance for SST-2 (ID) and fine-tuned OOD performance for CLINC-Banking, as shown in Table~\ref{tab:various-llms}. For simplicity, we report the AUROC result of each OOD score function. Overall, the OOD detection performance of LLaMA2-7B surpasses that of LLaMA-7B, and LLaMA-7B outperforms OPT-6.7B, which is consistent with the general performance trends observed in these models.

\subsection{Impact of Quantization}

\begin{table}[h]
\centering
\resizebox{\linewidth}{!}{
\begin{tabular}{c|cccc}
\hline
\hline
& \multicolumn{4}{c}{SST-2} \\
Quantized &  Maha & Cosine & MSP & Energy \\
\hline
float32  &  0.991 &  0.990  &  0.905  &  0.368  \\
float16 & 0.990 &  0.987  &  0.893  &  0.331 \\
Int8 &  0.955 &  0.960  &  0.874  &  0.228 \\
\hline
\hline
\end{tabular}
}
\caption{Impact of quantization for OOD detection.}
\label{tab:quantization}
\end{table}

We test the zero-grad OOD detection performance of LLaMA-7B with different quantized levels (float32, float16, and Int8) for SST-2 as the ID task, shown in Table~\ref{tab:quantization}. It was observed that the float16 model can largely preserve the OOD detection capability of the model. However, 8-bit quantization leads to a degradation in its capability.

\subsection{Error Analysis}

We mainly analyze the fine-tuning OOD detection performance in the near-OOD setting since both RoBERTa~\cite{liu2019roberta} and LLaMA~\cite{touvron2023llama} can achieve near-perfect performance in the far-OOD setting. Here, we provide the OOD detection performance with fine-tuned RoBERTa in the following Table~\ref{tab:error-analysis}, as a complement to Table~\ref{tab:fine_tune_llama}. 
\begin{table}[h]
\centering
\resizebox{\linewidth}{!}{
\begin{tabular}{c|cccc}
\hline
\hline
& \multicolumn{4}{c}{CLINC-Banking (Full)} \\
PTMs &  Maha & Cosine & MSP & Energy \\
\hline
LLaMA-7B &  \textbf{0.958} & \textbf{0.964} & \textbf{0.936} & \textbf{0.930}\\
RoBERTa-L & 0.821 &  0.793  &  0.670  &  0.717 \\
\hline
\hline
\end{tabular}
}
\caption{Performance comparison between LLaMA and RoBERTa-L in the near-OOD setting. AUROC result is reported for each score function.}
\label{tab:error-analysis}
\end{table}

Overall, LLaMA’s detection capabilities are much stronger than RoBERTa’s. Through the analysis of specific error cases, we found that RoBERTa struggles to distinguish semantically similar OOD samples. 
For instance, when choosing the ``bank\_balance'' class (e.g., the sentence ``what's my account balance'') as the ID distribution, RoBERTa tends to incorrectly classify the majority of        ``bill\_balance'' class inputs (such as ``what are my bills this month'') as ``bank\_balance''. In contrast, LLaMA generally makes correct judgments in most cases. 
We attribute this phenomenon to the anisotropy characteristic which is explained in Section~\ref{sec:analysis_generative}), i.e., sentence embeddings produced by the BERT family models have been noted to possess an undesirable characteristic of being concentrated within a narrow cone, causing representation degeneration. Despite this, in the cases involving extremely similar semantics, LLaMA also makes errors in judgment, such as misclassifying ``bill\_balance: what is the balance on my bills'' as ``bank\_balance''.

\section{Conclusion}
This paper has delved into the critical realm of OOD detection within the context of LLMs. The growing utilization of LLMs across various natural language processing tasks has underscored the need to understand their capabilities and limitations, especially in scenarios involving distribution shifts. 
Our work deepens the comprehension of OOD detection capabilities of LLMs. 
Through meticulous analysis, we have showcased the effectiveness of LLMs for OOD detection under various settings, including zero-grad and generative fine-tuning scenarios. 
Our findings reveal that a simple OOD detector utilizing the cosine similarity function outperforms other sophisticated OOD detectors, especially in the few-shot setting.
Our work may serve as a foundational stepping stone for future advancements in effectively and responsibly harnessing the potential of LLMs in diverse environments.

\textbf{Limitations and Future Work:}  
In this work, we only tested textual LLMs and ignored the exploration for multi-modal LLMs~\cite{liu2024visual}, which can be seen as a limitation. More OOD detection in multi-modal scenarios (including both general and medical domains)~\cite{liu2021slake, radford2021learning, zhu2023minigpt,liu2022medical,zhang2023llava} will be conducted in the future.

\section*{Acknowledgments}
We thank the anonymous reviewers for their valuable feedback. This research was partially supported by the grant of HK ITF ITS/359/21FP.


\section{Bibliographical References}\label{sec:reference}
\bibliographystyle{lrec-coling2024-natbib}
\bibliography{lrec-coling2024}


\end{document}